\documentclass[journal, twoside]{IEEEtran}
\usepackage{fancyhdr} 
\usepackage{lipsum}
\usepackage[utf8]{inputenc}
\usepackage[T1]{fontenc}
\usepackage{booktabs}
\usepackage{graphicx} 
\usepackage{subfig}
\usepackage{nccmath}
\usepackage{color}
\usepackage{cite}
\usepackage{acro}
\usepackage{scalerel}
\usepackage{tikz}
\usetikzlibrary{svg.path}
\usepackage{xspace}
\usepackage{amsmath} 
\usepackage{amsfonts} 
\usepackage{amssymb}
\usepackage{adjustbox}

\definecolor{orcidlogocol}{HTML}{A6CE39}
\tikzset{
  orcidlogo/.pic={
    \fill[orcidlogocol] svg{M256,128c0,70.7-57.3,128-128,128C57.3,256,0,198.7,0,128C0,57.3,57.3,0,128,0C198.7,0,256,57.3,256,128z};
    \fill[white] svg{M86.3,186.2H70.9V79.1h15.4v48.4V186.2z}
                 svg{M108.9,79.1h41.6c39.6,0,57,28.3,57,53.6c0,27.5-21.5,53.6-56.8,53.6h-41.8V79.1z M124.3,172.4h24.5c34.9,0,42.9-26.5,42.9-39.7c0-21.5-13.7-39.7-43.7-39.7h-23.7V172.4z}
                 svg{M88.7,56.8c0,5.5-4.5,10.1-10.1,10.1c-5.6,0-10.1-4.6-10.1-10.1c0-5.6,4.5-10.1,10.1-10.1C84.2,46.7,88.7,51.3,88.7,56.8z};
  }
}

\newcommand\orcidicon[1]{\href{https://orcid.org/#1}{\mbox{\scalerel*{
\begin{tikzpicture}[yscale=-1,transform shape]
\pic{orcidlogo};
\end{tikzpicture}
}{|}}}}

\usepackage{tikz}
\usetikzlibrary{shapes.geometric, arrows, positioning, fit, backgrounds}

\tikzstyle{startstop} = [rectangle, rounded corners, minimum width=3cm, minimum height=1cm,text centered, draw=black, fill=red!30]
\tikzstyle{process} = [rectangle, minimum width=3cm, minimum height=1cm, text centered, draw=black, fill=orange!30]
\tikzstyle{decision} = [diamond, aspect=2, minimum width=2cm, minimum height=1cm, text centered, draw=black, fill=green!30]
\tikzstyle{io} = [trapezium, trapezium left angle=70, trapezium right angle=110, minimum width=3cm, minimum height=1cm, text centered, draw=black, fill=blue!30]
\tikzstyle{arrow} = [thick,->,>=stealth]

\DeclareAcronym{acm}{
  short = ACM ,
  long  = Association for Computing Machinery ,
  sort  = A ,
}

\usepackage{hyperref} 
\begin{document}

 \setcounter{page}{1}

\title{Illustrating Classic Brazilian Books using a Text-To-Image Diffusion Model}

\author{Felipe Mahlow \orcidicon{0000-0001-9816-1440}\,, André Felipe Zanella \orcidicon{0000-0001-8828-6629}\,, William Alberto Cruz Castañeda \orcidicon{0000-0002-9803-1387}\,, 
        and~Regilene Aparecida Sarzi-Ribeiro  \orcidicon{0000-0001-6267-6549}\,
\thanks{F.M and R.A.S.R are with São Paulo State University (Unesp), Bauru Campus. A.F.Z is with Maringá State University, Maringá Campus. W.A.C.C. is with Technologycal Federal University of Paraná, Guarapuava Campus.}
\thanks{Manuscript received xx xx, 202x; revised xx xx, 202x.}}

\markboth{Mahlow et. al.: ILLUSTRATING CLASSIC BRAZILIAN BOOKS USING A TEXT-TO-IMAGE DIFFUSION MODEL.}{SKM: My IEEE article}  

\maketitle

\begin{abstract}
In recent years, Generative Artificial Intelligence (GenAI) has undergone a profound transformation in addressing intricate tasks involving diverse modalities such as textual, auditory, visual, and pictorial generation. Within this spectrum, text-to-image (TTI) models have emerged as a formidable approach to generating varied and aesthetically appealing compositions, spanning applications from artistic creation to realistic facial synthesis, and demonstrating significant advancements in computer vision, image processing, and multimodal tasks. The advent of Latent Diffusion Models (LDMs) signifies a paradigm shift in the domain of AI capabilities. This article delves into the feasibility of employing the Stable Diffusion LDM to illustrate literary works. For this exploration, seven classic Brazilian books have been selected as case studies. The objective is to ascertain the practicality of this endeavor and to evaluate the potential of Stable Diffusion in producing illustrations that augment and enrich the reader's experience. We will outline the beneficial aspects, such as the capacity to generate distinctive and contextually pertinent images, as well as the drawbacks, including any shortcomings in faithfully capturing the essence of intricate literary depictions. Through this study, we aim to provide a comprehensive assessment of the viability and efficacy of utilizing AI-generated illustrations in literary contexts, elucidating both the prospects and challenges encountered in this pioneering application of technology.
\end{abstract}


\IEEEpeerreviewmaketitle

\section{Introduction}
\label{intro}

Generative Artificial Intelligence (GenAI) has revolutionized various tasks by integrating capabilities in text, audio, video, and image generation. GenAI excels in creating synthetic data that closely mimics real-world phenomena. For instance, text generation models such as OpenAI's GPT \cite{openai2023gpt4} have transformed the field of writing by demonstrating an exceptional understanding of context and coherence \cite{roumeliotis2023chatgpt}. These models enhance natural language processing, content creation, and automated writing tasks \cite{huang2023role}. In the realm of audio, models like Tacotron \cite{wang2017tacotron} and WaveNet \cite{oord2016wavenet} leverage deep neural networks to generate realistic speech and music, pushing the boundaries of audio synthesis \cite{ning2019review}. Similarly, image generation has seen significant advancements with models such as DALL-E \cite{betker2023improving, ramesh2021zeroshot}, MidJourney \cite{midjourney}, and Stable Diffusion \cite{rombach2022highresolutionimagesynthesislatent}, which can create intricate images from textual descriptions. Additionally, Generative Adversarial Networks (GANs) \cite{goodfellow2016deep} have been pivotal in producing high-quality images for artistic endeavors and realistic face generation, significantly impacting computer vision and multi-modal tasks \cite{wang2018perceptual, liu2019multistage, qiao2019mirrorgan}. The ability of generative models to create human-like content has opened up new avenues for creative, automated, and innovative applications. However, these advancements also bring about concerns and challenges that need to be addressed in the future \cite{walczak2023challenges}. 

Text-to-image (TTI) models focus on developing methods and algorithms to create visual images from written text. A significant breakthrough in this field has been the rise of Latent Diffusion Models (LDMs) \cite{rombach2022highresolutionimagesynthesislatent}, which build upon the foundational principles of Diffusion Probabilistic Models (DPMs) \cite{ho2020denoising}. Diffusion models apply a series of random transformations progressively to an image's probability distribution \cite{croitoru2023diffusion}. This iterative process generates detailed and realistic images while providing greater control over the creation process.

The integration of GenAI into the creative process has shown a profound and multifaceted impact. For instance, research on the use of TTI models in craft education indicates that artificial intelligence (AI) can aid ideation and visualization but also raised concerns about skill gaps, lack of materiality, and ethical implications, including biases and the impact on creativity and copyright \cite{vartiainen2023using}. Similarly, GenAI can increase creativity in writing tasks, showing improved quality of text outputs with AI support, despite a greater uniformity in generated creations \cite{doshi2023generative}. Additionally, the application of AI in independent publishing, reveals both the AI’s capability to enhance ideation and production, and the need for critical approaches to its role in preserving human craftsmanship \cite{library993284}. AI use also has environmental implications, with research showing that text and illustration production with AI can generate significantly fewer carbon emissions compared to human methods \cite{tomlinson2024carbon}. However, the adoption of AI in creative fields is not without challenges, including ethical issues and the need for a careful balance between automation and human authorship. Analyzing these dynamics is crucial to understanding how AI can be a powerful tool in enhancing the creative process while addressing the challenges associated with its integration.

The extant literature offers a scant examination of the intersection between AI and book illustration. Traditionally, the illustration of literary works necessitated the intervention of human artists, a process often characterized by its time-consuming nature and subjectivity. However, recent advancements in AI present opportunities for automation, potentially enhancing this process by generating illustrations that encapsulate the essence and historical context of literary works with significantly reduced resources and time compared to traditional methodologies. An example of this is the GenAI methodology based on TTI for producing assisted art that allows dataset assembling, model training and fine-tuning, and content enhancement and post-processing within a historical or cultural setting \cite{genaimethodology}. Nevertheless, the endeavor of producing visually compelling and faithful images based on literary descriptions remains inherently complex. The quality and fidelity of these generated images are critically contingent upon the specificity of prompts and the efficacy of the models employed. Furthermore, challenges persist regarding data bias and the inherent limitations of AI models themselves. The training of these models is particularly contentious, as it often involves using copyrighted images without the explicit consent of the original artists \cite{10.1145/3597512.3597528}.

Our research explores the application of LDMs for book illustration, utilizing the Stable Diffusion model to generate visual representations based on textual prompts derived from seven distinct works of classic Brazilian literature. We employed a two-phase methodology that begins with initial image generation using Stable Diffusion XL Base 1.0, followed by a refinement process with Stable Diffusion XL Refiner 1.0. This iterative approach was designed to enhance the quality and fidelity of the generated illustrations, aligning them more closely with the literary descriptions. The aim of this work was not to achieve "perfect" illustrations but to demonstrate the feasibility of such an approach, providing a methodological framework and documenting the positive and negative aspects encountered, as well as the ethical biases inherent in GenAI.

Our findings underscore the critical importance of prompt specificity in generating high-quality visual content. By carefully designing concise prompts that capture the essence and context of the scenes described in the literature, we achieved accurate and engaging illustrations. However, our study also highlights challenges such as model biases, notably the tendency to produce predominantly white figures, as observed in the illustrations of characters from books like \textit{Senhora} and \textit{O Triste Fim de Policarpo Quaresma}. Quantitative evaluations using Contrastive Language-Image Pre-training (CLIP) and Inception Scores (IS) revealed varying success rates across different books, with better results achieved for works like \textit{Horto} and \textit{O Triste Fim de Policarpo Quaresma}, reflecting successful visualizations of the described concepts. These results offer valuable insights into the potential and limitations of using generative models for literary illustration and suggest areas for future improvements in prompt engineering and model training.

The paper is structured as follows: Section \ref{methods} outlines the methodology employed, starting with \ref{the_books}, where we present the books, followed by \ref{hardware}, where we discuss the hardware specifications and configurations used for training and inference. In \ref{image_gen}, we provide a detailed analysis of the image generation process, while in \ref{image_refinement} we address the techniques applied for refining these images. Further, in \ref{selection}, we detail the large-scale generation process and the subsequent selection of the most relevant images. Section \ref{metrics} is dedicated to the quantitative evaluation of the results, focusing on the CLIP Score and the IS, discussed in subsections \ref{clip} and \ref{inception_score}, respectively. Section \ref{results} presents an extensive exploration of the experimental results obtained, with a qualitative approach (\ref{qualitative}) illustrating the effectiveness and adaptability of our approach in generating diverse and visually appealing compositions in various contexts, as well as a quantitative approach (\ref{quantitative}) assessing the performance of the generated images based on established metrics. Finally, Section \ref{conclusions} synthesizes the main findings, highlights the effectiveness of the image synthesis pipeline, and discusses potential directions for future research and improvements to the proposed method.

\section{Methodology}
\label{methods}

This section describes the systematic procedure adopted for generating and refining illustrations for seven classical Brazilian books using the Stable Diffusion model. 

\subsection{The Books}
\label{the_books}

The selection of these texts was guided by the goal of respecting copyright constraints while providing a rich source for creative exercises. Therefore, only texts from Brazilian literature available in the public domain were chosen.

The selected books for this study are:

\begin{itemize}
    \item \textit{Senhora} (1875) by José de Alencar
    \item \textit{O Cortiço} (1890) by Aluísio Azevedo
    \item \textit{A Viúva Simões} (1897) by Júlia Lopes de Almeida
    \item \textit{Dom Casmurro} (1899) by Machado de Assis
    \item \textit{Horto} (1900) by Auta de Souza
    \item \textit{Os Sertões} (1902) by Euclides da Cunha
    \item \textit{O Triste Fim de Policarpo Quaresma} (1915) by Lima Barreto
\end{itemize}

These works were chosen for their literary significance and their rich descriptive passages, which provide ample material for visualization through GenAI. Each text offers unique narratives and vivid descriptions that facilitate the generation of diverse and engaging illustrations. \textit{Senhora} and \textit{O Cortiço}, for example, are notable for their detailed portrayal of 19th-century Brazilian society, while \textit{Dom Casmurro} and \textit{Os Sertões} are renowned for their deep psychological and socio-political insights. \textit{A Viúva Simões} and \textit{Horto} contribute with their unique thematic and stylistic elements, and \textit{O Triste Fim de Policarpo Quaresma} is recognized for its satirical and cultural critique. The choice of these works not only ensures a diverse range of illustrative challenges but also honors the literary heritage of Brazilian authors through modern technological means.

\subsection{Hardware Configuration}
\label{hardware}

All computations were performed on a system equipped with an NVIDIA GeForce RTX 3090 GPU. This setup provided enough computational power and memory capacity, essential for handling the process involved in generating and refining the images.

\subsection{Image Generation}
\label{image_gen}

In the first stage, the \textit{Stable Diffusion XL Base 1.0}\footnote{https://huggingface.co/stabilityai/stable-diffusion-xl-base-1.0} \cite{sdxl}  model was employed to generate the initial images. The model was fed with a comprehensive list of descriptive prompts, each crafted to depict specific scenes from the selected books. These prompts included detailed descriptions of characters, actions, and settings. For each prompt, the model performed the image generation in 40 inference steps. The inference process involved a denoising technique, crucial in diffusion models, which iteratively refined an image from an initial noisy state to a clear and detailed representation. In this case, the inference was configured to terminate at 0.8 of the denoising process, meaning that the image was generated to a certain level of detail before transitioning to the next stage.

\subsection{Image Refinement}
\label{image_refinement}

In the second stage, the \textit{Stable Diffusion XL Refiner 1.0}\footnote{https://huggingface.co/stabilityai/stable-diffusion-xl-refiner-1.0} model was used to enhance the images generated in the previous step. This refinement model received the latent images (partially processed images) and continued the denoising process from 0.8, completing the remaining steps to 1.0. This additional 40-step inference phase ensured that the final images reached a higher level of quality and detail. The refiner model utilized additional components, such as a second text encoder and the variational autoencoder (VAE) from the base model, to improve the accuracy and aesthetics of the images. The refiner corrected imperfections and added fine details, making the illustrations more vivid and true to the original descriptions.

\subsection{Large-Scale Generation and Selection}
\label{selection}

For each book, five prompts were generated, and for each prompt, 300 images, resulting in 1500 images per book. This large-scale generation ensured a wide variety of illustrations, allowing for a rigorous selection of the best representations for the project. These large quantities are important so that the calculation of quantitative metrics can be done more accurately. 

\subsection{Quantitative Evaluation Metrics}
\label{metrics}

To evaluate the quality of the images generated we employed two evaluation metrics, the CLIP \cite{hessel2022clipscorereferencefreeevaluationmetric} and IS scores \cite{salimans2016improvedtechniquestraininggans}.

\subsubsection{CLIP Score}
\label{clip}

The CLIP Score is a metric that evaluates the semantic quality and relevance of the generated images concerning the input prompts. We utilize the \textit{"clipq\_score"} function from the \textit{"torchmetrics.functional.multimodal"}\footnote{https://lightning.ai/docs/torchmetrics/stable/multimodal/clip\_score.html} library, which calculates the CLIP score of an image concerning a text prompt. The CLIP model was trained on a large number of text-image pairs to learn a joint representation of text and image. It is capable of assessing the semantic quality of the generated images concerning the provided prompts.

\subsubsection{Inception Score (IS)}
\label{inception_score}

IS is a metric that evaluates both the quality and diversity of the generated images. A higher IS indicates that the generated images are both diverse (capturing multiple classes) and of high quality (clearly recognizable by the classifier). The metrics were calculated through implementations based on the seminal paper\footnote{https://github.com/openai/improved-gan}\footnote{https://github.com/w86763777/pytorch-image-generation-metrics/blob/master/pytorch\_image\_generation\_metrics/inception.py}\cite{salimans2016improvedtechniquestraininggans}.

By employing these metrics, we ensure a comprehensive evaluation of the generated images, considering both their semantic relevance to the prompts and their overall quality and diversity.

\section{Results}
\label{results}

\begin{figure*}
    \centering
    \includegraphics[width=\linewidth]{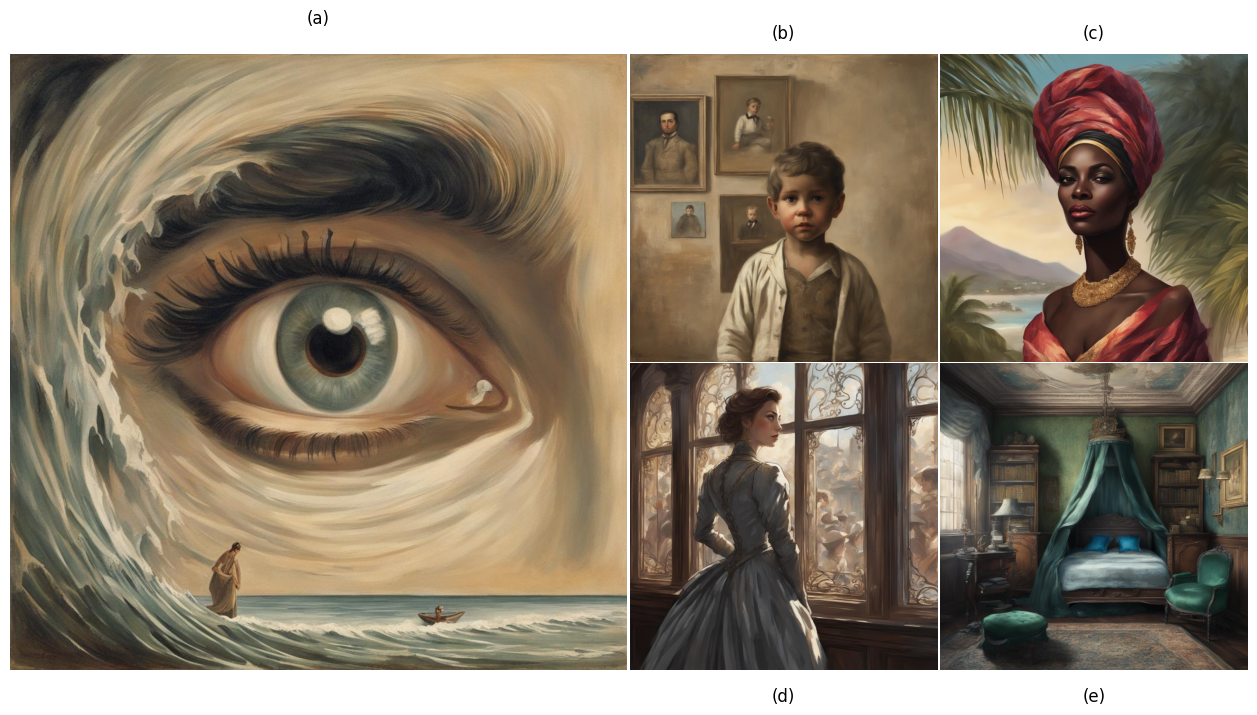}
    \caption{Examples of generated images for (a, b) \textit{Dom Casmurro}, (c) \textit{A Viúva Simões},  and (d, e) \textit{Senhora}.}
    \label{imgs1}
\end{figure*}

In this section, we present and analyze examples of the generated images, focusing on specific examples that highlight the process and outcomes of our methodology. Figures \ref{imgs1} and \ref{imgs2} showcase selected images, which facilitate a discussion on the creation process and the critical role of the prompt in generating visually captivating and contextually relevant illustrations.

While the AI system autonomously manages the image generation process, the formulation of the prompt remains a critical task for the user. Mere transcription of text from the book typically results in suboptimal outcomes. It is imperative to conceptualize the scene based on the textual descriptions provided in the book and devise a prompt that facilitates the creation of an image that is both visually compelling and faithful to the scene's intrinsic characteristics.

\subsection{Qualitative Approach}
\label{qualitative}

Figure \ref{imgs1}a) exemplifies the representation of Capitu's eyes, often described as "\textit{olhos de ressaca}" (sea surge eyes) in Machado de Assis's \textit{Dom Casmurro}. These eyes are characterized as "gypsy eyes, oblique and dissimulated," as mentioned in chapters 13 and 32 of the book. The prompts used for this image were inspired by the physical and symbolic traits detailed in these chapters. Capitu is described as having brunette hair and light eyes, considering that the true color of her eyes is not explicitly revealed in the book. The prompt used was: 

\begin{quote}
"Painting of oblique and concealed eyes. Just eyes. Mysterious and energetic fluid, like the wave that retreats from the beach, on hangover days. Brunette, clear and large eyes, straight and long nose, had a thin mouth and wide chin."
\end{quote}

Figure \ref{imgs1}b) clearly illustrates the importance of imagining a way to represent a scene. This image was created to depict Bentinho's suspicion that his son was fathered by Escobar, as described in chapters 131 and 132 of \textit{Dom Casmurro}, portraying Bentinho's obsession with Escobar's photograph, which he kept in his office. Bentinho frequently noted the resemblance between his son and Escobar, fueling his suspicions. The image positions the child in front of Escobar's photograph, highlighting the similarities between them and symbolizing Bentinho's growing distrust. The prompt used was:

\begin{quote}
"A painting of a young kid (4yo) in the center. In the background, there is a photograph of his father on the wall. The kid looks like the father, who was 40yo. 1800s."
\end{quote}

Figure \ref{imgs1}c) portrays the opulence of the character \textit{Ernestina}, based on \textit{Júlia Lopes de Almeida's} novel \textit{A Viúva Simões}. The image captures the essence of a rich widow, embodying her social status and physical attributes as described in chapter I of the book. The prompt used for this illustration was:

\begin{quote}
"Concept art of a rich woman, widow, forty years old, bourgeoisie, a beautiful woman, tall, slender, beautiful black eyes, dark skin that was delicately feathery and soft, Rio de Janeiro. Digital artwork, illustrative, painterly, matte painting, highly detailed."
\end{quote}

Figure \ref{imgs1}d) depicts \textit{Aurélia} at the window with her suitors, as described in part 2, chapter 4 of \textit{José de Alencar's} novel \textit{Senhora}. Despite the prompt not specifying skin color, all 300 images generated for "a young woman, very beautiful" depicted white women, indicating a bias in the algorithm. The image captures the essence of \textit{Aurélia's} character as she stands at the window, attracting numerous suitors. The prompt used for this illustration was:

\begin{quote}
"Concept art of a young woman, very beautiful, stands in front of the window. She attracts a crowd of suitors who pass by in carriages and on foot. The eager looks and insinuating words of the suitors contrast with her cold impassivity; she remains at the window like a statue, fulfilling her duty, but without employing flirtatious tricks or seduction tactics."
\end{quote}

Figure \ref{imgs1}e) illustrates \textit{Seixas's} room, highlighting the stark contrast between his luxurious lifestyle and the poverty of his family, as described in part 1, chapter 5 of \textit{José de Alencar's} novel \textit{Senhora}. The scene and specifically the room was chosen for illustration due to the detailed descriptions in the book, which reveal aspects of \textit{Seixas's} personality. This disparity is further emphasized as \textit{Seixas} changes marriages twice, driven by his desire to escape his circumstances and secure the dowry offered to him. The generated images, however, did not fully capture all the objects mentioned in the prompt, seemingly focusing primarily on the books. In this sense we emphasize that, to achieve optimal results, it is generally advisable to employ shorter and more succinct prompts that contain fewer elements within the image. The prompt used for this illustration was:

\begin{quote}
"Concept art of modest, worn-out study with faded blue wallpaper, old furniture. Iron bed with green mosquito net contrasts with surroundings. Luxurious items like a tailored black coat, elegant evening wear, a Parisian hat, quality gloves, and fine boots seem out of place. The embroidered blue satin pillow stands out. The disorderly alcove with books, inkwells, ashtrays, and assorted trinkets contrasts with the well-appointed dresser counter. Corner with umbrellas, canes, some valuable, alongside artistic curiosities."
\end{quote}

\begin{figure*}
    \centering
    \includegraphics[width=1\linewidth]{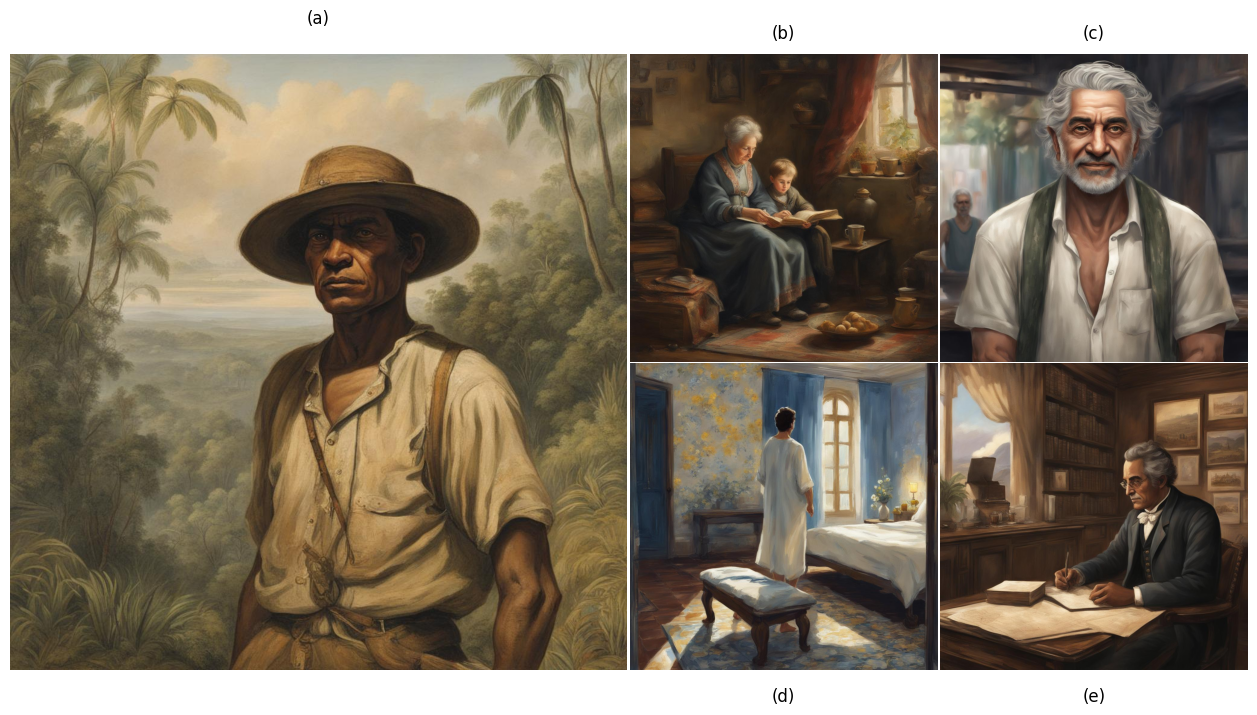}
    \caption{Examples of generated images for (a) \textit{Os Sertões}, (b) \textit{Horto}, (c) \textit{A Viúva Simões}, (d) O cortiço, and (e) \textit{O Triste Fim de Policarpo Quaresma}.}
    \label{imgs2}
\end{figure*}

Figure \ref{imgs2}a) portrays a man from the Brazilian backlands, reflecting his determined nature and the harsh life he endures, as described in part 2, chapter III of \textit{Euclides da Cunha's} \textit{Os Sertões}. The prompt creation was based on: \textit{"The sertanejo is, above all, a strong man. He does not have the exhaustive rickets of the neurasthenic mestizos of the coast. His appearance, however, at first glance, reveals the opposite. He lacks the impeccable physique, the poise, the correct structure of athletic organizations."}. Although the model faithfully represents the figure of the \textit{sertanejo}, all 300 generated images show dense vegetation in the background, not reflecting the true scenery of the Brazilian backlands. The prompt used for this illustration was:

\begin{quote}
"A painting of a man from the Brazilian backlands. He is a determined person, yet his countenance reflects the harsh life in the backlands, 19th century, realism."
\end{quote}

Figure \ref{imgs2}b) illustrates a scene inspired by \textit{Auta de Souza's} poem \textit{"À minha avó"} from her book \textit{Horto}. The generated image captures the warmth and comfort of the relationship between the grandmother and her grandson. The cozy environment, enhanced by elements such as a soft blanket, a cup of tea, and an old book, embodies the essence of comfort and security. However, it is important to note that the model also exhibited a bias towards generating white individuals, which is evident in the depiction of the characters. This particular book posed a challenge due to its abstract nature. The prompt used for the illustration was:

\begin{quote}
"A painting of a grandmother and her grandson (9yr) sitting together in a cozy environment, surrounded by elements that symbolize comfort and security, a soft blanket, a cup of tea, and an old book, 19th century, realism."
\end{quote}

Figure \ref{imgs2}c) illustrates the character \textit{Luciano}, described in chapters I and II, from \textit{Júlia Lopes de Almeida's} \textit{A Viúva Simões}. The generated image portrays a man with gray hair, a thick build, consistent with the description provided. His expressive and friendly face, marked by dark circles under his eyes, captures the intended character traits. This image serves as a clear example of the model's tendency to create more stylized, cartoon-like illustrations when guided by prompts that include "concept art." The prompt used for this illustration was:

\begin{quote}
"Concept art of a man, gray hair, thick, virile physiognomy, not slender, rounded belly, expressive and friendly face, dark circles under his eyes, Rio de Janeiro. Digital artwork, illustrative, painterly, matte painting, highly detailed."
\end{quote}

Figure \ref{imgs2}d) depicts a scene from \textit{Aluísio de Azevedo's} novel \textit{O Cortiço} chapter XXI. This depiction highlights the model's ability to capture the specific details and atmosphere described in the prompt. The highly detailed, painterly style aligns with the intent to represent the character \textit{João Romão}, a Portuguese owner of the collective housing, in a 19th-century Rio de Janeiro setting, showcasing the blend of period-appropriate elements with the everyday life of the character. The prompt used for this illustration was:

\begin{quote}
"Concept art of a man walking back in flip-flops and a nightgown in the bedroom, wide room lined in blue and white with little yellow flowers pretending to be gold, a rug at the foot of the bed, and on the bench a nickel alarm clock, Rio de Janeiro, 19th century. Digital artwork, illustrative, painterly, matte painting, highly detailed."
\end{quote}

Figure 2e) presents the character \textit{Policarpo Quaresma} from \textit{Lima Barreto's} \textit{O Triste Fim de Policarpo Quaresma}. The description comes from Chapter I and the image portrays a civil servant from the 19th century who is depicted as deeply valuing his country's culture. The illustration captures the essence of the character’s dedication and connection to the cultural heritage of Rio de Janeiro during that period. The prompt used for this illustration was:

\begin{quote}
"Concept art of a man, civil servant, values the country's culture, 19th century, Rio de Janeiro. Digital artwork, illustrative, painterly, matte painting, highly detailed."
\end{quote}

Broadly speaking, from an ethical standpoint, three primary issues highlighted in \cite{vartiainen2023using} are also immediately discernible in our work: misrepresentation (e.g., the harmful stereotyping of minority groups), underrepresentation (e.g., the eradication of the presence of a particular gender within specific professional roles), and overrepresentation (e.g., the predominance of Anglocentric viewpoints) \cite{bommasani2021opportunities}.

\subsection{Quantitative Approach}
\label{quantitative}

In this section, we present a quantitative analysis of the generated images using two evaluation metrics: CLIP and IS scores. These metrics provide insights into the semantic relevance and overall quality of the images produced for various literary concepts. Table \ref{tab:clip-is-evaluation} summarizes the evaluation results for each book concept. 

\begin{table}[h]
    \centering
    \begin{tabular}{@{}lcc@{}}
        \toprule
        \textbf{Concept} & \textbf{CLIP (std)} & \textbf{IS (std)} \\
        \midrule
        A Viúva Simões - Júlia Lopes & 20.44(1.63) & 5.47(0.22) \\
        Dom Casmurro - Machado de Assis & 17.96(1.79) & 5.91(0.32) \\
        Horto - Auta de Souza & 21.05(2.30) & 6.35(0.34) \\
        O Cortiço - Aluísio Azevedo & 20.11(2.41) & 3.76(0.23) \\
        O Triste Fim de P. Quaresma - L. Barreto & 18.94(2.23) & 7.87(0.36) \\
        Os Sertões - Euclides da Cunha & 18.71(1.55) & 6.20(0.45) \\
        Senhora - José de Alencar & 20.28(1.33) & 3.86(0.23) \\
        \bottomrule
    \end{tabular}
    \caption{Evaluation results using the CLIP and IS metrics for each concept.}
    \label{tab:clip-is-evaluation}
\end{table}

It should be observed that, despite the limited presentation of images — restricted to no more than two distinct prompts per book in Figures \ref{imgs1} and \ref{imgs2} — each book is characterized by five distinct prompts with 300 images each. The aforementioned metrics are thus computed based on an aggregate of 1500 generated images. The results indicate varying levels of performance across different book concepts. For the CLIP Score, \textit{Horto} by \textit{Auta de Souza} achieved the highest score of 21.05 (standard deviation 2.30), reflecting strong semantic alignment between the generated images and the text prompts. This is followed by \textit{A Viúva Simões} with a CLIP Score of 20.44 (standard deviation 1.63) and \textit{Senhora} by \textit{José de Alencar} with a score of 20.28 (standard deviation 1.33). On the lower end, \textit{Dom Casmurro} scored 17.96 (standard deviation 1.79), indicating comparatively weaker semantic relevance.

In terms of the IS, which assesses the image quality and diversity, \textit{O Triste Fim de Policarpo Quaresma} achieved the highest score of 7.87 (standard deviation 0.36), suggesting superior image quality and variability. \textit{Horto} follows with a score of 6.35 (standard deviation 0.34), indicating high-quality results but with less diversity compared to the previous. Conversely, \textit{O Cortiço} and \textit{Senhora} exhibited lower scores, 3.76 (standard deviation 0.23) and 3.86 (standard deviation 0.23) respectively, pointing to challenges in achieving high image quality and diversity.

Overall, these quantitative results highlight differences in the effectiveness of the generative models across various literary texts, reflecting the impact of both semantic alignment and image quality on the final outputs.

\section{Conclusions}
\label{conclusions}

In this study, we explored the application of LDMs for the task of book illustration, utilizing the Stable Diffusion model to generate images based on prompts derived from seven classical Brazilian literary works. Our results indicate that the effectiveness of image generation is significantly influenced by the quality and specificity of the prompts provided. Prompts that were carefully crafted to encapsulate the essence of the scenes described in the books yielded more compelling and relevant visual results. Conversely, generic or poorly articulated prompts often led to suboptimal outcomes, highlighting the importance of prompt design in the generative process.

Through the illustrative examples provided, including figures representing characters and scenes from \textit{Dom Casmurro}, \textit{A Viúva Simões}, \textit{Senhora}, \textit{O Cortiço}, \textit{O Triste Fim de Policarpo Quaresma}, and \textit{Os Sertões}, we observed that the Stable Diffusion model could effectively capture and convey the thematic and visual elements of the source material. However, certain limitations were noted. For instance, the model's tendency to produce predominantly white individuals, despite the diversity described in the books, points to inherent biases in the training data that affect the generated results. This is exemplified by the generated images of characters from \textit{Senhora} and \textit{O Triste Fim de Policarpo Quaresma}. 

Quantitative evaluations using CLIP and IS scores revealed variations in the quality of the generated images across different concepts. The analysis demonstrated that images aligned with literary descriptions from \textit{Horto} and \textit{O Triste Fim de Policarpo Quaresma} scored high in both metrics, indicating successful visualizations of these concepts.

In conclusion, while the Stable Diffusion model is a powerful tool for generating illustrative content from textual prompts, its efficacy relies on precise prompt formulation and awareness of model biases. Future models should address these issues by incorporating more diverse datasets. AI research in illustration should also focus on advanced prompt engineering to enhance the quality and inclusivity of generated images. This study adds to the growing body of research on using generative models in creative contexts, offering insights into their potential and limitations in literary illustration.
\section*{Acknowledgments}
F. M. acknowledges support from Coordena\c{c}{\~a}o de Aperfei\c{c}oamento de Pessoal de N{\'i}vel Superior (CAPES), project number 88887.607339/2021-00. A. F. Z. acknowledges support from Conselho Nacional de Desenvolvimento Científico e Tecnológico (CNPq), project number 140935/2024-0.

\bibliographystyle{IEEEtran}  

\bibliography{References}

\vspace{-1.5cm}
\begin{IEEEbiography}[{\includegraphics[width=1in,height=1.25in,clip,keepaspectratio]{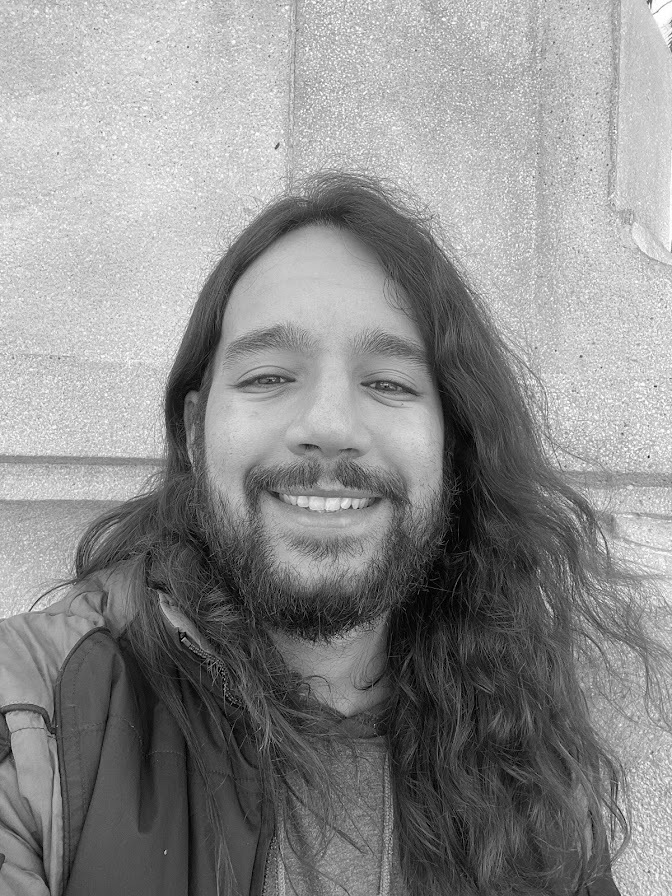}}]{1st author}
received a teaching degree in Physics and a bachelor's degree in Materials Physics from São Paulo State University (Unesp), in 2020 and 2023, respectively. He is currently pursuing a Ph.D. degree in Computer Science at São Paulo State University (Unesp), with a focus on classical and quantum machine learning and their applications to Quantum Information Science. His research encompasses the use of Machine Learning and Generative Artificial Intelligence, as well as Quantum Computing.
\end{IEEEbiography}
\vspace{-1.5cm}

\begin{IEEEbiography}[{
\includegraphics[width=1in,height=1.25in,clip,keepaspectratio]{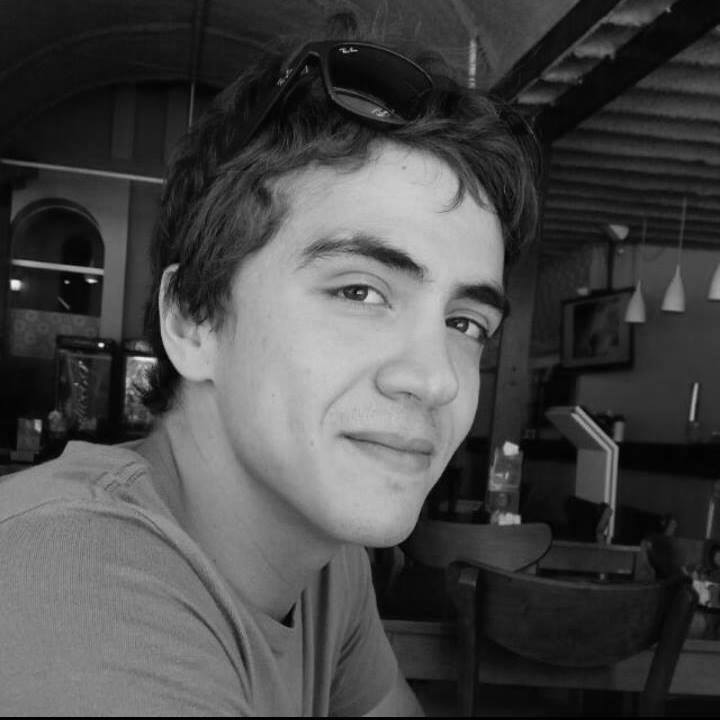}
}]
{2nd author}
Holds a Bachelor's degree in Mathematics (2022) and a Master's degree in Computer Science (2022), both from the State University of Maringá. Currently, he is pursuing a Ph.D. in Computer Science at the same institution (2024). His expertise lies in Computer Science, focusing on Computer Systems and Machine Learning. He is interested in research in optimization, machine learning, and TTI diffusion models.
\end{IEEEbiography}
\vspace{-1.5cm}

\begin{IEEEbiography}[{\includegraphics[width=1in,height=1.25in,clip,keepaspectratio]{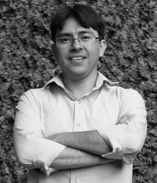}}]
{3rd author}
Researcher in AI holds a Bachelor's degree in Computer Science from Benemérita Universidad Autónoma de Puebla and a Bachelor's degree in Computer Engineering from the Federal University of Rio Grande do Sul. He completed his Master's and Ph.D. in Electrical Engineering with a focus on Biomedical Engineering at the Federal University of Santa Catarina. He holds a Postdoctoral degree in AI and Biomedical Engineering at the State University of Santa Catarina. Currently, he is a professor at the Federal Technological University of Paraná, Guarapuava Campus. His research interests include ubiquitous computing, machine learning, and GenAI.
\vspace{-2cm}

\end{IEEEbiography}
\begin{IEEEbiography}[{\includegraphics[width=1in,height=1.25in,clip,keepaspectratio]{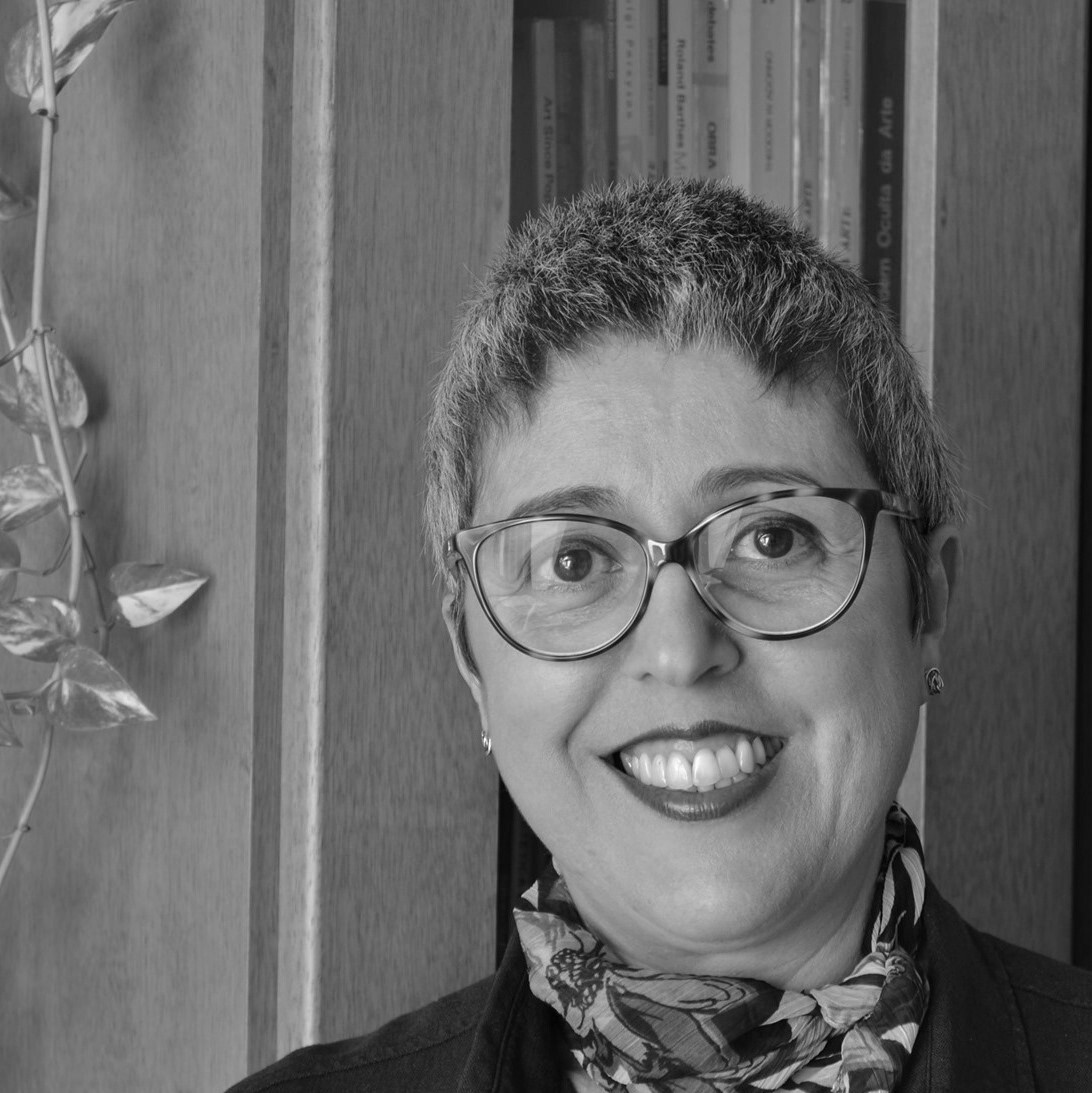}}]
{4th author}
Holds Postdoctoral degrees in Poetics and Cultures in Digital Humanities at UFG (2022) and Arts at UNESP/SP (2013). She is part of the Chair of Design, Art, and Science at Media Lab BR and coordinates the Media Lab UNESP. Research member of Red de Investigación de la Imagen at the University of Málaga, Spain. Holds a Ph.D. in Communication and Semiotics from PUC/SP (2012) and a Master's in Arts from UNESP/SP (2007). Graduated in Art Education with a specialization in Visual Arts from FAAC - UNESP, Bauru/SP (1994). Currently, faculty member at UNESP/Bauru, coordinates the Graduate Program in Media and Technology, and leads the labIMAGEM Research Group.

\end{IEEEbiography}
\end{document}